\documentclass[sigconf]{acmart}
\usepackage{booktabs}
\usepackage{multirow}
\usepackage{subcaption} 

\usepackage{bm}

\usepackage{algorithm}
\usepackage{algorithmic} 

\usepackage{enumitem}
\usepackage{listings}
\usepackage{balance}

\AtBeginDocument{%
  }

\copyrightyear{2026}
\acmYear{2026}
\acmConference[WWW '26]{Proceedings of the ACM Web Conference 2026}{April 13--17, 2026}{Dubai, United Arab Emirates}
\acmBooktitle{Proceedings of the ACM Web Conference 2026 (WWW '26), April 13--17, 2026, Dubai, United Arab Emirates}
\acmPrice{}
\acmDOI{10.1145/3774904.3792463}
\acmISBN{979-8-4007-2307-0/2026/04}

\begin{document}

\title{RaDAR: Relation-aware Diffusion-Asymmetric Graph Contrastive Learning for Recommendation}

\author{Yixuan Huang}
\orcid{0009-0005-5854-6668}
\authornote{These authors contributed equally to this work.}
\affiliation{%
  \institution{University of Electronic Science and Technology of China}
  \city{Chengdu}
  \country{China}
}
\email{wangdaijiayou@163.com}

\author{Jiawei Chen}
\orcid{0009-0004-4236-6235}
\authornotemark[1]
\affiliation{%
  \institution{National University of Defense Technology}
  \city{Changsha}
  \country{China}
}
\email{cjw@nudt.edu.cn}

\author{Shengfan Zhang}
\orcid{0009-0000-5074-3492}
\affiliation{%
  \institution{Ant Group}
  \city{Chongqing}
  \country{China}
}
\email{15060126965@163.com}

\author{Zongsheng Cao}
\orcid{0009-0006-5607-3999}
\authornote{Corresponding author.}
\affiliation{%
  \institution{Tsinghua University}
  \city{Beijing}
  \country{China}
}
\email{agiczsr@gmail.com}

\renewcommand{\shortauthors}{Yixuan Huang, Jiawei Chen, Shengfan Zhang, \& Zongsheng Cao}

\begin{abstract}
Collaborative filtering (CF) recommendation has been significantly advanced by integrating Graph Neural Networks (GNNs) and Graph Contrastive Learning (GCL). However, (i) random edge perturbations often distort critical structural signals and degrade semantic consistency across augmented views, and (ii) data sparsity hampers the propagation of collaborative signals, limiting generalization. To tackle these challenges, we propose \textbf{RaDAR} (\textbf{R}elation-\allowbreak\textbf{a}ware \textbf{D}iffusion-\allowbreak\textbf{A}symmetric 
\textbf{Graph Contrastive Learning Framework for \textbf{R}ecommendation \textbf{Systems}}), a novel framework that combines two complementary view generation mechanisms: a graph generative model to capture global structure and a relation‑aware denoising model to refine noisy edges. RaDAR introduces three key innovations: (1) \emph{asymmetric contrastive learning} with global negative sampling to maintain semantic alignment while suppressing noise; (2) \emph{diffusion‑guided augmentation}, which employs progressive noise injection and denoising for enhanced robustness; and (3) \emph{relation‑aware edge refinement}, dynamically adjusting edge weights based on latent node semantics. Extensive experiments on three public benchmarks demonstrate that RaDAR consistently outperforms state‑of‑the‑art methods, particularly under noisy and sparse conditions.
Our code is available at our repository\footnote{\url{https://github.com/ohowandanliao/RaDAR}}.
\end{abstract}

\ccsdesc[500]{Information systems~Recommender systems}
\ccsdesc[500]{Mathematics of computing~Graph algorithms}

\keywords{Recommendation, Diffusion Model, Contrastive Learning, Data Augmentation}


\maketitle

\section{Introduction}
With the development of Artificial Intelligence\cite{cao2025cofi,cao2025purifygen,cao2025tv}, recommender systems\cite{wu2022graph} play a vital role in alleviating information overload by learning personalized preferences from sparse user-item interactions. A prevailing approach to recommendation is collaborative filtering (CF)\cite{schafer2007collaborative}, which infers user interests based on historical behavioral patterns. To capture high-order connectivity and structural semantics, recent methods have leveraged Graph Neural Networks (GNNs) \cite{scarselli2008graph}, which model user-item interactions through message passing on bipartite graphs. These advances have significantly improved recommendation accuracy, particularly in sparse settings.

To further enhance representation learning, Graph Contrastive Learning (GCL)\cite{you2020graph} has emerged as a self-supervised paradigm that encourages consistency across multiple augmented views of the interaction graph. By integrating GCL with GNNs, recent models aim to improve robustness against data sparsity and noise. Typical implementations, such as SGL~\cite{wu2021self}, generate graph augmentations through node or edge dropout, while methods like GraphACL~\cite{xiao2023simple} introduce asymmetric contrastive objectives to capture multi-hop patterns. In parallel, diffusion-based models~\cite{ho2020denoising,cao2024diffusione, li2022diffusion} have shown promise in improving denoising capacity through iterative noise injection and reconstruction.

\begin{figure}[t]  
    \centering
    \includegraphics[width=0.5\textwidth]{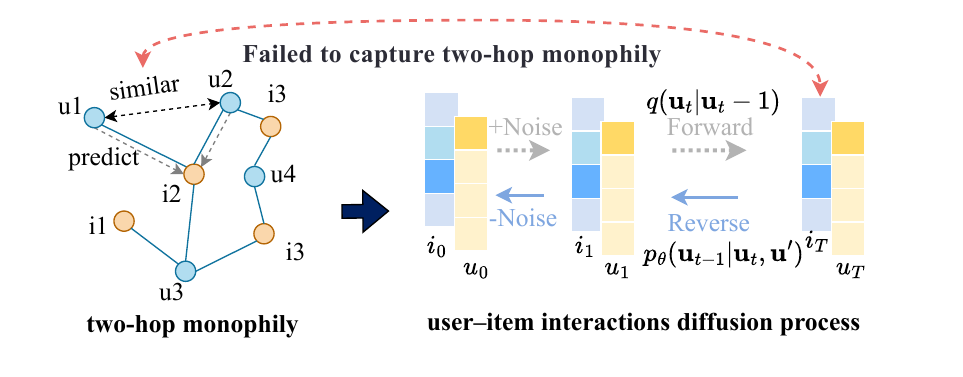}    
    \caption{ACL and diffusion model on a user–item graph, illustrating how standard diffusion misses two-hop monophily where indirectly connected users share similar preferences.}
    \label{fig:intro}
\end{figure}

Despite these advancements, two fundamental challenges limit current GCL-based recommendation models:
\textbf{Challenge 1 (C1): Structural Semantics Degradation.}
Standard graph augmentations (e.g., random node/edge dropout) often corrupt essential topological structures, degrading collaborative signals and destabilizing contrastive learning. This structural perturbation compromises semantic consistency between augmented views, hindering effective representation learning.
\textbf{Challenge 2 (C2): Limited Relational Expressiveness.}
Existing methods predominantly assume homophily, emphasizing one-hop neighborhood alignment. However, real-world user interactions frequently exhibit heterophily or distant homophily—where similar users connect through multi-hop paths with weak direct links. Current models inadequately capture these higher-order relational patterns. While diffusion models enhance noise robustness, they sacrifice fine-grained relational semantics beyond immediate neighborhoods.
As illustrated in Fig.~\ref{fig:intro}, two-hop neighbors often share implicit preferences despite weak direct connections, which are not captured by conventional approaches. 
This raises a key question: \textit{How can we design a unified model that preserves structural semantics during augmentation while learning relation-aware representations across multi-hop and heterophilous neighborhoods?}

To address recommendation challenges in sparse and noisy scenarios, we propose \textbf{RaDAR} (\underline{R}elation-aware \underline{D}iffusion-\underline{A}symmetric Graph Contrastive Learning for \underline{R}ecommendation), a contrastive learning framework with two core objectives: preserving structural semantics and enhancing relational expressiveness.

For \textbf{C1 (structural semantics degradation)}, RaDAR introduces a \emph{diffusion-guided augmentation strategy} applying Gaussian noise to node representations with learned denoising. This maintains semantic integrity while generating robust graph views for contrastive learning, reducing overfitting to spurious patterns.

For \textbf{C2 (limited relational expressiveness)}, RaDAR employs a \emph{dual-view generation architecture} combining: (i) a graph generative module based on variational autoencoders, capturing global structural semantics beyond one-hop connections; and (ii) a relation-aware graph denoising module that adaptively reweights edge contributions, preserving fine-grained relational signals. Additionally, RaDAR's \emph{asymmetric contrastive objective} decouples node identity from structural context, enabling alignment of semantically similar nodes even in heterogeneous neighborhoods.

Comprehensive experiments on both binary-edge and weighted-edge benchmarks—covering public datasets 
(Last.FM, Yelp, BeerAdvocate) and multi-behavior datasets (Tmall, RetailRocket, IJCAI15)—demonstrate 
that \textbf{RaDAR} consistently outperforms state-of-the-art baselines. The model achieves particularly strong 
gains under high sparsity and noise conditions, demonstrating its robustness and adaptability across interaction regimes.

In summary, our main contributions are threefold:
\begin{itemize}[leftmargin=*, itemsep=1pt]
    \item We present \textbf{RaDAR}, a dual-view contrastive framework that integrates diffusion-guided augmentation with relation-aware denoising for robust representation learning;
    \item We design an asymmetric contrastive objective that enhances structural discrimination while mitigating noise via DDR-style diffusion regularization;
    \item RaDAR achieves consistent state-of-the-art performance across both binary-edge and weighted-edge recommendation settings, demonstrating strong generalization and noise resilience.
\end{itemize}

\section{Preliminaries and Related Work}

\subsection{Collaborative Filtering Paradigm}
Let $ U $ and $ V $ denote user and item sets, with interactions encoded in a binary matrix. Graph-based collaborative filtering extracts representations by propagating information across the interaction graph under the homophily principle: users with similar interaction patterns share preferences.
Implementations typically employ dual-tower architectures to map users and items into a shared latent space, enabling relevance estimation through similarity matching. This approach captures transitive dependencies in interaction graphs to infer unobserved user-item affinities.

\begin{figure*}[ht]
    \centering
    \includegraphics[width=\textwidth]{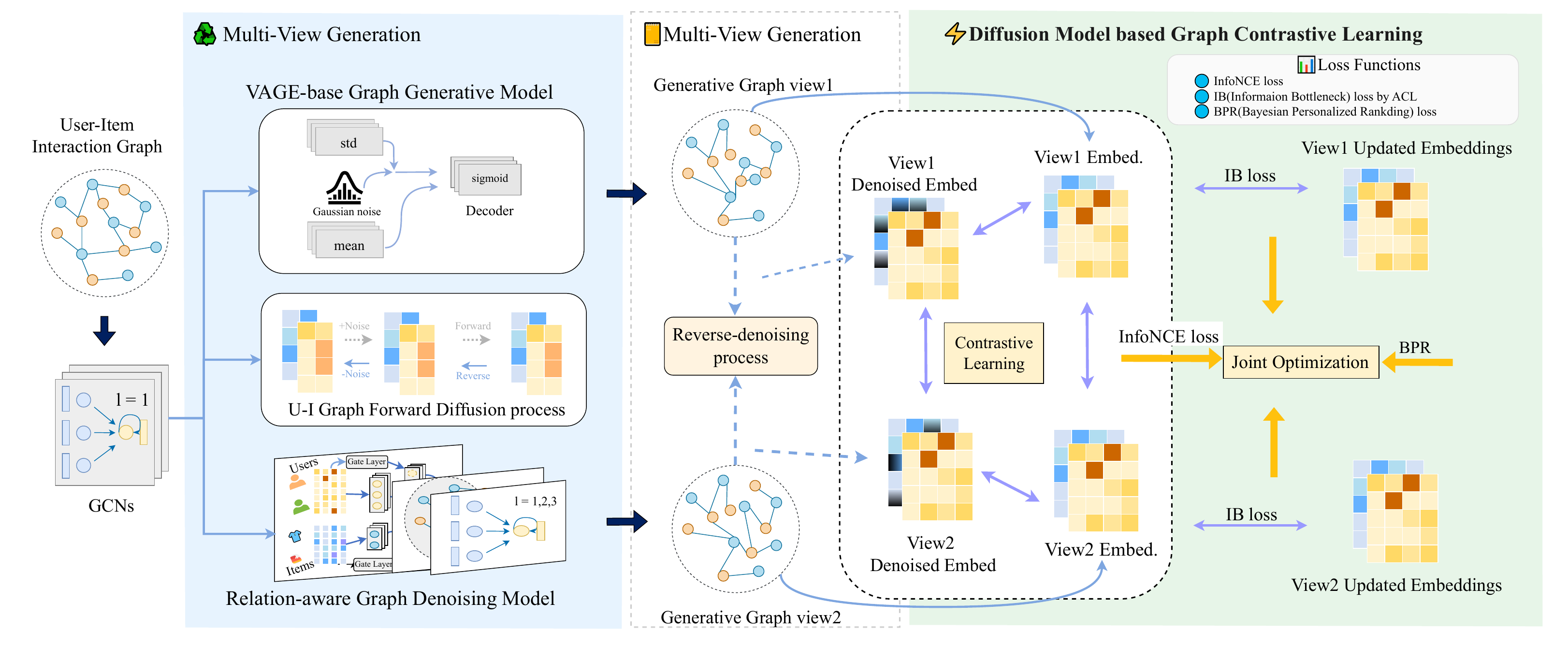}    
    \caption{RaDAR framework architecture: The left section shows two view generators extracting complementary graph representations. The right section demonstrates the contrastive learning process with diffusion model-based graph generation and joint optimization through InfoNCE, IB, and BPR losses.}
    \label{fig:framework}
\end{figure*}
\subsection{Self-Supervised Graph Learning}

Recent advances in graph neural networks (GNNs) have transformed recommendation systems by explicitly modeling user–item relations. 
Core architectures such as PinSage~\cite{ying2018graph}, NGCF~\cite{wang2019neural}, and LightGCN~\cite{he2020lightgcn} 
encode multi-hop relational patterns through graph convolutions, with LightGCN simplifying propagation for higher efficiency. 
Subsequent works enhance representation learning via multi-intent disentanglement (DGCF~\cite{wang2020disentangled}, DCCF~\cite{ren2023disentangled}) 
and adaptive relation discovery (DRAN~\cite{wang2022learning}). 
Temporal dynamics are further captured by sequence-aware GNNs (DGSR~\cite{zhang2022dynamic}, GCE-GNN~\cite{wang2020global}) 
that connect historical interactions with evolving preferences.

The fusion of self-supervised learning (SSL) and graph modeling has become a leading paradigm for data-efficient representation learning. 
Contrastive methods (e.g., SGL~\cite{wu2021self}, GFormer~\cite{li2023graph}) improve user–item embeddings through view-invariance, 
while reconstruction-based models (S3-Rec~\cite{zhou2020s3}) employ masked interaction prediction. 
Recent studies extend SSL to cross-domain and multi-modal settings (C2DSR~\cite{cao2022contrastive}, SLMRec~\cite{tao2022self}), 
highlighting its versatility in leveraging auxiliary self-supervision. 

\noindent \textbf{Positioning and Scope.} 
RaDAR lies within the collaborative filtering (CF) paradigm on bipartite user–item graphs, 
and is designed to operate consistently under both binary- and weighted-edge regimes. 
Unlike models that rely on large-scale pretraining or side information, 
RaDAR unifies two complementary view-generation strategies, generative reconstruction and relation-aware denoising, within a single training pipeline. 
This unified design generalizes across edge settings via a simple weight toggle, 
minimizing implementation variance and ensuring controlled, fair evaluation in subsequent experiments.

\section{Methodology}
In this section, we present the comprehensive architecture of RaDAR, comprising four interconnected components. 
The first component employs a graph message passing encoder to effectively capture local collaborative relationships between users and items. The second component implements a sophisticated user-item graph diffusion model. The third component integrates an adaptive framework featuring two distinct trainable view generators: one that leverages a graph variational model and another that utilizes relation-aware denoising graph models. The fourth component focuses on model optimization through a multi-faceted loss function that incorporates ACL to boost performance, complemented by diffusion model-based Graph Contrastive Learning. The overall architecture of the RaDAR model is illustrated in Figure \ref{fig:framework}.

\subsection{User-item Embedding Propagation}
We project users and items into a $d$-dimensional latent space through learnable embeddings, denoted as $\mathbf{E}^{(u)} \in \mathbb{R}^{N \times d}$ and $\mathbf{E}^{(v)} \in \mathbb{R}^{M \times d}$ for $N$ users and $M$ items. To capture collaborative signals, we employ a normalized adjacency matrix derived from the interaction matrix:
\begin{equation}
    \label{eq:norm_adj}
    \tilde{\boldsymbol{A}} = \boldsymbol{D}_u^{-\frac{1}{2}} \boldsymbol{A} \boldsymbol{D}_v^{-\frac{1}{2}}
\end{equation}
where $\boldsymbol{D}_u$ and $\boldsymbol{D}_v$ are diagonal degree matrices for users and items.

The embedding propagation process utilizes a multi-layer graph neural network where user and item representations are iteratively refined through message passing:
\begin{equation}
    \label{eq:gnn_prop}
    \begin{aligned}
        \mathbf{E}_l^{(u)} &= \tilde{\mathbf{A}} \mathbf{E}_{l-1}^{(v)} + \mathbf{E}_{l-1}^{(u)} \\
        \mathbf{E}_l^{(v)} &= \tilde{\mathbf{A}}^\top \mathbf{E}_{l-1}^{(u)} + \mathbf{E}_{l-1}^{(v)}
    \end{aligned}
\end{equation}

The final embeddings integrate information across all $L$ layers through summation:
\begin{equation}
    \label{eq:final_emb}
    \mathbf{E}^{(u)} = \sum_{l=0}^L \mathbf{E}_l^{(u)}, \quad \mathbf{E}^{(v)} = \sum_{l=0}^L \mathbf{E}_l^{(v)}
\end{equation}

We compute the preference score between user $u_i$ and item $v_j$ via the inner product of their respective embeddings:
\begin{equation}
    \label{eq:pref_score}
    \hat{y}_{i,j}=(e_i^{(u)})^\top e_j^{(v)}
\end{equation}

\subsection{GCL Paradigm}
\subsubsection{\bfseries Graph Generative Model as View Generator}
We adopt Variational Graph Auto-Encoder (VGAE) \cite{kipf2016variational} for view generation, integrating variational inference with graph reconstruction. The encoder employs multi-layer GCN for node embeddings, while the decoder reconstructs graph structures using Gaussian-sampled embeddings.
The VGAE framework optimizes a multi-component loss function comprising KL-divergence regularization (Eq.~\ref{eq:kl_div}), discriminative loss for reconstructing graph structure (Eq.~\ref{eq:vgae_dis}), and Bayesian Personalized Ranking loss (Eq.~\ref{eq:bpr_loss}). The complete formulation of the VGAE objective is provided in Appendix~\ref{appendix:vgae_details} (Eq.~\ref{eq:gen_loss}).

\subsubsection{\bfseries Relation-Aware Graph Denoising for View Generation}

Our denoising framework employs a layer-wise edge masking strategy with sparsity constraints to generate clean graph views. The core idea is to model edge retention probabilities through reparameterized Bernoulli distributions, where the parameters are learned via relation-aware denoising layers.

The layer-wise edge masking is formulated as:
\begin{equation}
    \begin{aligned}
    A^l &= A \odot M^l,\\
    \sum_{l=1}^L |M^l|_0 &= \sum_{l=1}^L \sum_{(u,v)\in \epsilon} \mathbb{I}(m^l_{u,v} \neq 0)
    \end{aligned}
\label{eq:sparsity}
\end{equation}
where $A^l$ denotes the masked adjacency matrix at layer $l$, $M^l$ is the binary mask matrix, and $\tau_{\text{sparse}}$ controls the overall sparsity budget across all layers.

To preserve essential user-item relationships while filtering noise, our denoising layer employs adaptive gating mechanisms:
\begin{equation}
    \begin{aligned}
    \mathbf{g} &= \sigma(\mathbf{W}_g[\mathbf{e}_i;\mathbf{e}_j] + \mathbf{b}) \\
    \alpha_{i,j}^l &= f_{\text{att}}\left(\mathbf{G}(\mathbf{e}_i,\mathbf{e}_j) \oplus \mathbf{G}(\mathbf{e}_j,\mathbf{e}_i) \oplus [\mathbf{e}_i;\mathbf{e}_j]\right)
    \end{aligned}
    \label{eq:gating}
\end{equation}
where $\mathbf{g}$ represents the adaptive gate vector, and $\alpha_{i,j}^l$ denotes the attention weight for edge $(i,j)$ at layer $l$. The adaptive feature composition $\mathbf{G}(\cdot,\cdot)$ combines relational context with node embeddings:
\begin{equation}
    \mathbf{G}(\mathbf{e}_i,\mathbf{e}_j) = \mathbf{g} \odot \tau(\mathbf{W}_{\text{embed}}[\mathbf{e}_i;\mathbf{a}_{r,i}]) + (1-\mathbf{g})\odot\mathbf{e}_i \label{eq:adaptive_composition}
\end{equation}
where $\mathbf{a}_{r,i}$ represents the relational feature vector for node $i$. The framework utilizes a GRU-inspired mechanism \cite{cho2014learning} for relational filtering and employs a concrete distribution for differentiable edge sampling. The edge sampling process uses a concrete distribution with a hard sigmoid rectification to enable end-to-end training:
\begin{equation}
    \mathcal{L}_c = \sum_{l=1}^{L} \sum_{(u_i, v_j) \in \epsilon} \left( 1 - \mathbb{P}\sigma(s_{i,j}^l \mid \theta^l) \right) \label{eq:concrete_dist}
\end{equation}
where $s_{i,j}^l$ represents the edge score and $\theta^l$ are the learnable parameters for layer $l$. The training objective combines concrete distribution regularization with recommendation loss:
\begin{equation}
    \mathcal{L}_{\text{den}} = \mathcal{L}_c + \mathcal{L}_{\text{bpr}}^{\text{gen}} + \lambda_2\|\Theta\|_{F}^{2}
    \label{eq:final_loss}
\end{equation}
where $\mathcal{L}_{\text{bpr}}^{\text{gen}}$ is the BPR loss computed on the denoised graph views, and the last term provides L2 regularization on model parameters $\Theta$.

\subsection{Diffusion with User-Item Graph} Building on diffusion models' noise-to-data generation capabilities\cite{wang2023diffusion, ho2020denoising, sohl2015deep}, we propose a graph diffusion framework that transforms the original user-item graph $\mathcal{G}_{ui}$ into recommendation-optimized subgraphs $\mathcal{G}_{ui}'$. We design a forward-inverse diffusion mechanism: forward noise injection gradually degrades node embeddings via Gaussian perturbations, while inverse denoising recovers semantic patterns through learned transitions. This process enhances robustness against interaction noise while learning complex embedding distributions. The restored embeddings produce probability distributions for subgraph reconstruction, establishing an effective diffusion paradigm for high-fidelity recommendation graph generation.

\subsubsection{\textbf{Noise Diffusion Process}}

Our framework introduces a latent diffusion paradigm for graph representation learning, operating on GCN-derived embeddings rather than graph structures. Let $\mathbf{h}^{L}$ denote the item embedding from the final GCN layer. We construct a $T$-step Markov chain $\bm{\chi}_{0:T}$ with initial state $\bm{\chi}_{0}=\mathbf{h}_0^{(L)}$.

The forward process progressively adds Gaussian noise to embeddings, transforming them towards a standard normal distribution. Through reparameterization techniques (detailed in Appendix A.1), we can directly compute any intermediate state from the initial embedding:
\begin{equation}
    \bm{\chi}_{t} = \sqrt{\bar{\alpha}_t} \bm{\chi}_{0} + \sqrt{1-\bar{\alpha}_t}\bm{\epsilon}, \bm{\epsilon} \sim \mathcal{N}(0, \textbf{I})
\end{equation}
To precisely control noise injection, we implement a linear noise scheduler with hyperparameters $s$, $\alpha_{low}$, and $\alpha_{up}$ Appendix~\ref{app:DetailedDiffusionProcessFormulation}.

The reverse process employs neural networks parameterized by $\theta$ to progressively denoise representations, recovering the original embeddings through learned Gaussian transitions. This denoising procedure enables our model to capture complex patterns in the graph-derived embeddings while maintaining their structural properties.

\subsubsection{\bf Diffusion Process Optimization for User-Item Interaction.}
The optimization objective is formulated to maximize the Evidence Lower Bound (ELBO) of the item embedding likelihood $\bm{\chi}_0$. Following the diffusion framework in \cite{jiang2024diffkg}, we derive the training objective as:
\begin{equation}
    \mathcal{L}_{elbo} = \mathbb{E}_{t\sim\mathcal{U}(1,\textit{T})}\mathcal{L}_t.
\end{equation}
where $\mathcal{L}_{t}$ denotes the loss at diffusion step $t$, computed by uniformly sampling timesteps during training. 
The ELBO comprises two components: (1) a reconstruction term $\mathbb{E}_{q(\bm{\chi}_{1}|\bm{\chi}_{0})}\left[\lVert\hat{\bm{\chi}}_{\theta}(\bm{\chi}_{1},1)-\bm{\chi}_{0}\rVert_{2}^{2}\right]$ 
that evaluates the model's denoising capability at $t=1$, and (2) KL regularization terms governing the reverse process transitions.
Following \cite{jiang2024diffkg}, we minimize the KL divergence between the learned reverse distribution
$p_\theta(\bm{\chi}_{t-1}|\bm{\chi}_{t})$ and the tractable posterior $q(\bm{\chi}_{t-1}|\bm{\chi}_{t}$
The neural network $\hat{\bm{\chi}}_{\theta}(\cdot)$, implemented as a Multi-Layer Perceptron (MLP), predicts the original embedding 
$\bm{\chi}_{0}$ from noisy embeddings $\bm{\chi}_{t}$ and timestep encodings. This formulation preserves the theoretical guarantees of ELBO maximization.

\subsection{Contrastive Learning paradigms}
\subsubsection{Diffusion-Enhanced Graph Contrastive Learning}\label{ssec:diff_cl}
We propose a diffusion-augmented contrastive framework leveraging intra-node self-discrimination for self-supervised learning. Given node embeddings $E^{'}$ and $E^{''}$ from two augmented views, we consider augmented views of the same node as positive pairs ($e'_i, e''_i$), and views of different nodes as negative pairs ($e'_i, e''_{i'}$) where $u_i \neq u_{i'}$. 
The formulation of the loss function is:
\begin{equation}
    \mathcal{L}_{ssl}^{user} = \sum_{u_i \in \mathcal{U}} -\log \frac{\exp(s(e_i', e_i'') / \tau)}{\sum_{u_j \in \mathcal{U}} \exp(s(e_i', e_j'') / \tau)}, \label{eq:ssl_user_loss}
\end{equation}
where $s(\cdot)$ denotes cosine similarity and $\tau$ represents the temperature parameter. The item-side contrastive loss $\mathcal{L}_{ssl}^{item}$ follows an analogous formulation. The complete self-supervised objective combines both components:
\begin{equation}
    \mathcal{L}_{ssl} = \mathcal{L}_{ssl}^{user} + \mathcal{L}_{ssl}^{item} \label{eq:ssl_loss}
\end{equation}

Our diffusion-enhanced augmentation generates denoised views ($\mathcal{V}_1^{den}$, $\mathcal{V}_2^{den}$) via Markov chains that preserve interaction patterns while suppressing high-frequency noise. The framework implements:
(i) \textit{Intra-View Alignment} ($L_{\text{intra}}$), which measures the contrastive loss between original view $\mathcal{V}_i$ and its denoised counterpart $\mathcal{V}_i^{den}$.
(ii) \textit{Inter-View Regularization} ($L_{\text{inter}}$), which computes the contrastive loss between different denoised views $\mathcal{V}_1^{den}$ and $\mathcal{V}_2^{den}$.

The composite loss integrates these mechanisms:
\begin{equation}
    L_{\text{diff-ssl}} = L_{\text{ssl}} + \lambda_1 L_{\text{intra}} + \lambda_2 L_{\text{inter}} \label{eq:diff_ssl_loss}
\end{equation}
where $\lambda_1$ and $\lambda_2$ balance view consistency and information diversity. This design enables simultaneous noise suppression and multi-perspective representation learning.

\textbf{Diffusion–ACL Synergy.}
Diffusion and asymmetric contrastive learning (ACL) play complementary roles in enhancing representation quality. 
Diffusion smooths high-frequency noise and refines local neighborhoods, stabilizing latent embeddings. 
ACL aligns structurally similar nodes across multi-hop relations by decoupling identity from context. 
Their combination enables RaDAR to suppress noise while preserving monophily-style semantics beyond one hop. 
Empirical results (Sec.~\ref{sec:complexity}, RQ1/RQ2) confirm that diffusion-augmented ACL consistently improves NDCG with comparable or better Recall.

\subsubsection{\textbf{Asymmetric Graph Contrastive Learning}}\label{ssec:asym_cl}
Conventional contrastive frameworks are limited by homophily assumptions \cite{lim2021large, chin2019decoupled}. We adopt an asymmetric paradigm \cite{xiao2023simple} for monophily-structural contexts using dual encoders $f_{\theta}$ and $f_{\xi}$ that generate identity and context representations. An asymmetric predictor reconstructs neighborhood contexts from node identities (Eq.~\ref{eq:acl_loss} in Appendix~\ref{sec::acl}). This preserves node semantics while encoding structural patterns, naturally accommodating monophily through shared central nodes.
Our dual-representation framework uses view-specific encoders $f_{\theta}$ and $f_{\xi}$ to generate identity representations $\mathbf{v} = f_{\theta}(G)[\mathbf{v}]$ and context representations $\mathbf{u} = f_{\xi}(G)[\mathbf{u}]$. An asymmetric predictor $g_{\phi}$ reconstructs neighborhood contexts from node identities, optimizing a contrastive objective (see Eq.~\ref{eq:acl_loss} in Appendix~\ref{sec::acl}).
 
This formulation achieves two key properties: (1) identity representations preserve node-specific semantics, and (2) context representations encode structural neighborhood patterns. The asymmetric objective naturally accommodates monophily by enabling two-hop neighbors to reconstruct similar contexts through their shared central nodes.
 
\subsection{Model Training}
Our framework adopts a hierarchical optimization approach with three coupled stages, as summarized in Table \ref{tab:training_phases}.

\textbf{Phase 1: Unified Multi-Task Learning} We initiate joint optimization:
\begin{equation}
    L_{\text{1}} = L_{\text{bpr}} + \lambda_3 L_{\text{diff-ssl}} + \lambda_4 \|\Theta\|_F^2
\end{equation}
where $L_{\text{diff-ssl}}$ is the diffusion-based self-supervised loss from Eq.~\ref{eq:diff_ssl_loss}, and $\|\Theta\|_F^2$ is L2 regularization.

\textbf{Phase 2: Representation Distillation}
We impose an information bottleneck constraint:
\begin{equation}
    \begin{aligned}
        \mathcal{L}_{IB} &= L_{A}(G, g_{\phi}(\mathbf{v}), \mathbf{v}, \mathbf{u}) \\[-.5ex]
                             &= L_{A}(G, g_{\phi}(\mathbf{y^*}), \mathbf{y^*}, \mathbf{\hat{y}}), \label{eq:ib_loss}
    \end{aligned}
\end{equation}
where \( \mathbf{y^*} \) represents historical representations and $L_{A}$ is the ACL loss.

\textbf{Phase 3: View Generator Optimization}
We finalize training by optimizing view generators:
\begin{equation}
    \mathcal{L}_{generators} = \mathcal{L}_{gen} + \mathcal{L}_{den}
\end{equation}
where $\mathcal{L}_{gen}$ is the VGAE graph generation loss(see Eq.~\ref{eq:gen_loss}) and $\mathcal{L}_{den}$(see Eq.~\ref{eq:final_loss}) is the relation-aware denoising loss.

\begin{table}[t]
\caption{Training phases in our framework.}
\label{tab:training_phases}
\centering
\scalebox{1.0}{
    \begin{tabular}{c|l|l}
        \hline
        \textbf{Phase} & \textbf{Objective} & \textbf{Params} \\
        \hline
        1 & $L_{\text{bpr}} ,L_{\text{diff-ssl}}, \|\Theta\|_F^2$ & User-item embeds \\
        \hline
        2 & $\mathcal{L}_{IB}$ (Info. bottleneck) & User-item embeds \\
        \hline
        3 & $\mathcal{L}_{gen} + \mathcal{L}_{den}$ & View generators \\
        \hline
        \end{tabular}
}
\end{table}

\subsubsection{\textbf{DDR-Style Denoising Warmup}}
Following DDRM~\cite{wang2023diffusion}, a lightweight diffusion regularizer is optionally applied to item embeddings after a brief warmup to enhance ranking robustness and avoid early-stage drift, consistent with DDRM’s delayed diffusion supervision.

\begin{table}[t]
    \centering
    \caption{Statistics of weighted-edge datasets.}
    \small
    \setlength{\tabcolsep}{4pt}
    \scalebox{0.90}{
        \begin{tabular*}{\linewidth}{@{\extracolsep{\fill}}l|c|c|c|l@{}}
            \toprule
            Dataset & Users & Items & Links & Interaction Types \\ \midrule
            Tmall        & 31,882 & 31,232 & 1,451,29 & View, Favorite, Cart, Purchase \\
            RetailRocket & 2,174  & 30,113 & 97,381   & View, Cart, Transaction \\
            IJCAI15      & 17,435 & 35,920 & 799,368  & View, Favorite, Cart, Purchase \\
            \bottomrule
        \end{tabular*}
    }
    \label{tab:weighted_stats}
\end{table}

\begin{table}[t]
    \centering
    \caption{Statistics of the experimental datasets.}
    \label{tab:dataset_stats_binary}
    \small
    \setlength{\tabcolsep}{4pt}
    \scalebox{0.90}{
        \begin{tabular*}{\linewidth}{@{\extracolsep{\fill}}l|c|c|c|c@{}}
            \toprule
            Dataset & Users & Items & Interactions & Density \\ 
            \midrule
            Last.FM      & 1,892   & 17,632  & 92,834         & $2.8 \times 10^{-3}$ \\
            Yelp         & 42,712  & 26,822  & 182,357        & $1.6 \times 10^{-4}$ \\
            BeerAdvocate & 10,456  & 13,845  & 1,381,094      & $9.5 \times 10^{-3}$ \\
            \bottomrule
        \end{tabular*}
    }
\end{table}

\begin{table*}[h]
    \centering
    \caption{Performance Metrics for Various Models}
    \label{tab:baseline}
    \resizebox{\textwidth}{!}{  
        \begin{tabular}{@{}|l|c|c|c|c|c|c|c|c|c|c|c|c|@{}}
            \hline
            \multirow{2}{*}{Model} & \multicolumn{4}{c|}{Last.FM} & \multicolumn{4}{c|}{Yelp} & \multicolumn{4}{c|}{BeerAdvocate} \\
            \cline{2-13}
            & Recall@20 & NDCG@20 & Recall@40 & NDCG@40 & Recall@20 & NDCG@20 & Recall@40 & NDCG@40 & Recall@20 & NDCG@20 & Recall@40 & NDCG@40 \\
            \hline
            BiasMF    & 0.1879 & 0.1362 & 0.2660 & 0.1653 & 0.0532 & 0.0264 & 0.0802 & 0.0321 & 0.0996 & 0.0856 & 0.1602 & 0.1016 \\
            NCF       & 0.1130 & 0.0795 & 0.1693 & 0.0952 & 0.0304 & 0.0143 & 0.0487 & 0.0187 & 0.0729 & 0.0654 & 0.1203 & 0.0754 \\
            AutoR     & 0.1518 & 0.1114 & 0.2174 & 0.1336 & 0.0491 & 0.0222 & 0.0692 & 0.0268 & 0.0816 & 0.0650 & 0.1325 & 0.0794 \\
            PinSage   & 0.1690 & 0.1228 & 0.2402 & 0.1472 & 0.0510 & 0.0245 & 0.0743 & 0.0315 & 0.0930 & 0.0816 & 0.1553 & 0.0980 \\
            STGCN     & 0.2067 & 0.1558 & 0.2940 & 0.1821 & 0.0562 & 0.0282 & 0.0856 & 0.0355 & 0.1003 & 0.0852 & 0.1650 & 0.1031 \\
            GCMC      & 0.2218 & 0.1714 & 0.3149 & 0.1897 & 0.0584 & 0.0280 & 0.0891 & 0.0360 & 0.1082 & 0.0901 & 0.1766 & 0.1085 \\
            NGCF      & 0.2081 & 0.1474 & 0.2944 & 0.1829 & 0.0681 & 0.0336 & 0.1019 & 0.0419 & 0.1033 & 0.0873 & 0.1653 & 0.1032 \\
            GCCF      & 0.2222 & 0.1642 & 0.3083 & 0.1931 & 0.0724 & 0.0365 & 0.1151 & 0.0466 & 0.1035 & 0.0901 & 0.1662 & 0.1062 \\
            LightGCN  & 0.2349 & 0.1704 & 0.3220 & 0.2022 & 0.0761 & 0.0373 & 0.1175 & 0.0474 & 0.1102 & 0.0943 & 0.1757 & 0.1113 \\
            SLRec     & 0.1957 & 0.1442 & 0.2792 & 0.1737 & 0.0665 & 0.0327 & 0.1032 & 0.0418 & 0.1048 & 0.0881 & 0.1723 & 0.1068 \\
            NCL       & 0.2353 & 0.1715 & 0.3252 & 0.2033 & 0.0806 & 0.0402 & 0.1230 & 0.0505 & 0.1131 & 0.0971 & 0.1819 & 0.1150 \\
            SGL       & 0.2427 & 0.1761 & 0.3405 & 0.2104 & 0.0803 & 0.0398 & 0.1226 & 0.0502 & 0.1138 & 0.0959 & 0.1776 & 0.1122 \\
            HCCF      & 0.2410 & 0.1773 & 0.3232 & 0.2051 & 0.0789 & 0.0391 & 0.1210 & 0.0492 & 0.1156 & 0.0990 & 0.1847 & 0.1176 \\
            SHT       & 0.2420 & 0.1770 & 0.3235 & 0.2055 & 0.0794 & 0.0395 & 0.1217 & 0.0497 & 0.1150 & 0.0977 & 0.1799 & 0.1156 \\
            DirectAU  & 0.2422 & 0.1727 & 0.3356 & 0.2042 & 0.0818 & 0.0424 & 0.1226 & 0.0524 & 0.1182 & 0.0981 & 0.1797 & 0.1139 \\
            AdaGCL    & \underline{0.2603} & \underline{0.1911} & \underline{0.3531} & \underline{0.2204} & \underline{0.0873} & \underline{0.0439} & \underline{0.1315} & \underline{0.0548} & \underline{0.1216} & \underline{0.1015} & \underline{0.1867} & \underline{0.1182} \\
            \textbf{Ours}      & \textbf{0.2724} & \textbf{0.1992} & \textbf{0.3664} & \textbf{0.2309} & \textbf{0.0914} & \textbf{0.0464} & \textbf{0.1355} & \textbf{0.0571} & \textbf{0.1262} & \textbf{0.1056} & \textbf{0.1946} & \textbf{0.1375} \\
            \hline
            Improv       & 4.65\% & 4.24\% & 3.77\% & 4.76\% & 4.70\% & 5.69\% & 3.04\% & 4.20\% & 3.78\% & 4.04\% & 4.23\% & 16.33\% \\
            \hline
            p-val    &$2.4e^{-6}$ &$5.8e^{-5}$ &$4.9e^{-9}$ &$6.4e^{-5}$ &$1.3e^{-4}$ &$8.8e^{-4}$ &$7.6e^{-3}$ &$2.2e^{-3}$ &$1.2e^{-4}$ &$7.9e^{-4}$ &$1.4e^{-4}$ &$2.9e^{-6}$ \\
            \hline
        \end{tabular}
    }
\end{table*}

\section{Experimental Evaluation}
To rigorously evaluate the proposed model, we design experiments to investigate four critical aspects:

\begin{itemize}[leftmargin=*]
    \item \textbf{RQ1}: How does \textit{RaDAR} perform against state-of-the-art recommendation baselines in benchmark comparisons?
    \item \textbf{RQ2}: What is the individual contribution of key components to the model's effectiveness across diverse datasets? (Ablation Analysis)
    \item \textbf{RQ3}: How robust is \textit{RaDAR} in handling data sparsity and noise compared to conventional approaches?
    \item \textbf{RQ4}: How do critical hyperparameters influence the model's performance characteristics?
\end{itemize}

\subsection{Model Complexity Analysis}
\label{sec:complexity}
We summarize the dominant costs on a bipartite graph with $|\mathcal{U}|$ users, $|\mathcal{V}|$ items, $|\mathcal{E}|$ edges, embedding size $d$, and $L$ propagation layers.

\paragraph{Time Complexity.}
Per epoch, RaDAR has three components:
\begin{itemize}[leftmargin=*]
    \item \textbf{Sparse propagation:} $L$ GCN-style message passing costs 
    $\mathcal{O}(|\mathcal{E}|\,d\,L)$.
    \item \textbf{Contrastive objective:} the implementation computes all-pairs similarities within a mini-batch (users+items, size $B$), yielding 
    $\mathcal{O}(B^{2} d)$ per step.
    \item \textbf{Diffusion regularization (optional):} one denoising pass on item embeddings costs 
    $\mathcal{O}(|\mathcal{V}|\, d d')$ (or $\mathcal{O}(S B d')$ if step-based), typically smaller than propagation when $|\mathcal{E}|$ is large.
\end{itemize}
Overall, the training is dominated by sparse propagation; contrastive and diffusion add a modest, data/mini-batch dependent overhead.

\paragraph{Memory Complexity.}
We store user/item embeddings and $L$ layer outputs: $\mathcal{O}((|\mathcal{U}|+|\mathcal{V}|)d)$. The diffusion MLP contributes $\mathcal{O}(d^{2}+ d d')$ parameters, independent of graph size. Hence the footprint remains comparable to lightweight CF GCNs.

\subsection{Experimental Settings}

\subsubsection{\textbf{Evaluation Datasets}}
We evaluate our method on three publicly available datasets:
\begin{itemize}[leftmargin=*]
\item \textbf{Last.FM}\cite{cantador2011second}: Music listening behaviors and social interactions from Last.fm users.
\item \textbf{Yelp} \cite{YelpDataset}: A benchmark dataset of user-business ratings from Yelp, widely utilized in location-based recommendation studies.
\item \textbf{BeerAdvocate} \cite{mcauley2013amateurs}: Beer reviews from BeerAdvocate, preprocessed with 10-core filtering to ensure data density.
\end{itemize}

Additionally, we adopt three widely used \emph{weighted-edge} (multi-behavior) e-commerce datasets: \textbf{Tmall}\cite{tmall_tianchi}, \textbf{RetailRocket}\cite{retailrocket}, and \textbf{IJCAI15}\cite{ijcai15_tianchi}. Dataset statistics are summarized in Table~\ref{tab:dataset_stats_binary} (binary) and Table~\ref{tab:weighted_stats} (weighted). Following public protocols (e.g., DiffGraph/HEC-GCN), we merge multiple behaviors into a single interaction graph and treat edges as binary presence; graph propagation uses the standard symmetric normalized adjacency.

\noindent \textbf{Binary vs. Weighted-Edge Regimes.} 
RaDAR is evaluated under two complementary settings: 
binary edges capture implicit feedback presence, 
whereas the weighted-edge setting follows a multi-behavior protocol that aggregates behaviors into a single binary interaction graph. 
This dual setup enables fair assessment of edge denoising on sparse graphs and robustness in heterogeneous e-commerce data. 
aligning our evaluation with recent multi-behavior studies such as DiffGraph~\cite{li2025diffgraph}.

\subsubsection{\textbf{Evaluation Protocols}} Following standard evaluation protocols for recommendation systems, we partition datasets into training/validation/test sets (7:2:1). Adopting the all-ranking strategy, we evaluate each user by ranking all non-interacted items alongside test positives. Performance is measured using Recall@20 and NDCG@20 metrics, with $K=20$ as the default ranking cutoff. This setup ensures a comprehensive assessment of model capabilities in real-world sparse interaction scenarios. For the weighted-edge (multi-behavior) regime, we follow the public construction and evaluation protocol: behaviors are aggregated into a binary interaction graph and evaluated with the same all-ranking pipeline.

\subsubsection{\textbf{Compared Baseline Methods}}
We evaluate RaDAR against a comprehensive suite of representative recommendation models spanning both single- and multi-behavior paradigms. 
Specifically, we categorize the baselines into four major research streams: 
(1) \textbf{Traditional collaborative filtering}: BiasMF~\cite{koren2009matrix} and NCF~\cite{he2017neural}; 
(2) \textbf{GNN-based methods}: LightGCN~\cite{he2020lightgcn} and NGCF~\cite{wang2019neural}; 
(3) \textbf{Self-supervised frameworks}: SGL~\cite{wu2021self} and SLRec~\cite{yao2021self}; 
and (4) \textbf{Contrastive and diffusion learning}: DirectAU~\cite{wang2022towards} and AdaGCL~\cite{jiang2023adaptive}. 

For the \emph{weighted-edge} regime, following DiffGraph’s protocol~\cite{li2025diffgraph}, 
we further include multi-behavior and heterogeneous baselines widely adopted in e-commerce scenarios, 
including BPR~\cite{koren2009matrix}, PinSage~\cite{ying2018graph}, NGCF~\cite{wang2019neural}, NMTR~\cite{gao2019neural}, 
MBGCN~\cite{jin2020multi}, HGT~\cite{hu2020heterogeneous}, and MATN~\cite{xia2020multiplex}. 
We also incorporate DDRM~\cite{wang2023diffusion} for completeness as a recent diffusion-based generative recommender.

\noindent \textbf{Binary vs. Weighted Evaluation Narrative.} 
We first report results under the binary-edge setting to benchmark each model’s collaborative filtering and contrastive learning capability without behavioral intensities. 
We then extend the evaluation to the weighted-edge setting, which introduces interaction-level heterogeneity to assess robustness under multi-behavior conditions. 
This order highlights that RaDAR’s improvements are not tied to a specific edge construction but arise from its unified design—integrating view generation and diffusion-based asymmetric contrastive learning (ACL)—that transfers seamlessly across regimes.

To avoid redundancy, shared baselines (e.g., BPR, PinSage, NGCF) are evaluated under both settings but described only once. 
Full baseline descriptions and implementation details are provided in Appendix~\ref{app:baselines}. 
This taxonomy provides comprehensive coverage from foundational to state-of-the-art paradigms, enabling a rigorous and balanced evaluation across methodological dimensions.

\subsection{Overall Performance (RQ1)}
Table~\ref{tab:baseline} summarizes the results on the three binary-edge benchmarks. 
RaDAR achieves the best performance across all datasets and cutoffs (top-20/40), 
surpassing strong CF, GNN, and SSL baselines with statistically significant improvements (see Improv and p-val rows). 
These gains verify that relation-aware denoising and diffusion-guided augmentation jointly enhance both coverage and ranking quality while preserving structural semantics.

For the weighted-edge regime, following DiffGraph’s protocol, 
Table~\ref{tab:weighted_edge_single} reports results on Tmall, RetailRocket, and IJCAI15. 
Under the same training pipeline (enabling edge weights), 
RaDAR consistently outperforms public multi-behavior baselines and the reproduced DiffGraph across all datasets. 
The DDR variant further improves NDCG@20, reflecting its emphasis on ranking robustness in heterogeneous interactions.

\begin{table}[t]
\centering
\caption{Weighted-edge results under DiffGraph settings (Recall@20 / NDCG@20).}
\label{tab:weighted_edge_single}
\small
\setlength{\tabcolsep}{2.5pt}
\resizebox{0.48\textwidth}{!}{
\begin{tabular}{l|cc|cc|cc}
\toprule
\multirow{2}{*}{Model} & \multicolumn{2}{c|}{Tmall} & \multicolumn{2}{c|}{RetailRocket} & \multicolumn{2}{c}{IJCAI15} \\
 & R@20 & N@20 & R@20 & N@20 & R@20 & N@20 \\
\midrule
BPR        & 0.0248 & 0.0131 & 0.0308 & 0.0237 & 0.0051 & 0.0037 \\
PinSage    & 0.0368 & 0.0156 & 0.0423 & 0.0248 & 0.0101 & 0.0041 \\
NGCF       & 0.0399 & 0.0169 & 0.0405 & 0.0257 & 0.0091 & 0.0035 \\
NMTR       & 0.0441 & 0.0192 & 0.0460 & 0.0265 & 0.0108 & 0.0048 \\
MBGCN      & 0.0419 & 0.0179 & 0.0492 & 0.0258 & 0.0112 & 0.0045 \\
HGT        & 0.0431 & 0.0192 & 0.0413 & 0.0250 & 0.0126 & 0.0051 \\
MATN       & 0.0463 & 0.0197 & 0.0524 & 0.0302 & 0.0136 & 0.0054 \\
DiffGraph  & 0.0553 & 0.0254 & 0.0626 & 0.0353 & 0.0178 & 0.0067 \\
\midrule
RaDAR (w)         & \textbf{0.0626} & \textbf{0.0268} & \textbf{0.1380} & \textbf{0.0746} & 0.0582 & 0.0323 \\
RaDAR (w+DDR)     & 0.0620 & 0.0260 & 0.1375 & 0.0748 & \textbf{0.0603} & \textbf{0.0325} \\
\bottomrule
\end{tabular}
}
\end{table}

\subsection{Model Ablation Test (RQ2)}
\begin{table}
    \centering
    \caption{Ablation study on key components of RaDAR.}
    \label{tab:ablation}
    \scalebox{0.80}{
        \begin{tabular}{l|c|cc|cc|cc}
            \hline\hline
            \multirow{2}{*}{Model} & Variant & \multicolumn{2}{c|}{Last.FM} & \multicolumn{2}{c|}{Yelp} & \multicolumn{2}{c}{Beer} \\
            & Description & Recall & NDCG   & Recall & NDCG   & Recall & NDCG   \\
            \hline
            Baseline                  & SOTA SSL    & 0.2603 & 0.1911 & 0.0873 & 0.0439 & 0.1216 & 0.1015 \\
            \hline
            \multirow{4}{*}{RaDAR} & Gen+Gen     & 0.2665 & 0.1936 & 0.0900 & 0.0456 & 0.1226 & 0.1027 \\
            & Gen+Linear  & 0.2698 & 0.1986 & 0.0910 & 0.0461 & 0.1247 & 0.1050 \\
            & w/o D-ACL     & 0.2652 & 0.1934 & 0.0904 & 0.0458 & 0.1250 & 0.1036 \\
             & w/ ACL only  & 0.2720 & 0.1962 & 0.0911 & 0.0461 & 0.1264 & 0.1057 \\
            \hline
            \multicolumn{2}{c|}{RaDAR(full)} & 0.2724 & 0.1992 & 0.0914 & 0.0464 & 0.1262 & 0.1056 \\
            \hline\hline
        \end{tabular}
    }
\end{table}
To evaluate RaDAR's architectural components, we conducted systematic ablation studies against the state-of-the-art baseline. We examined four configurations across three datasets (Last.FM, Yelp, and Beer):
\begin{itemize}[leftmargin=*, itemsep=0pt]
\item \textbf{RaDAR (Gen+Gen):} Dual VGAE-based generators without denoising model
\item \textbf{RaDAR (Gen+Linear):} Linear attention replacing relation-aware denoising model
\item \textbf{RaDAR (w/o D-ACL):} Conventional graph contrastive loss without diffusion-asymmetric contrastive learning optimization
\item \textbf{RaDAR (w/ ACL only):} Asymmetric contrastive learning without diffusion-based augmentation
\item \textbf{RaDAR (full):} Complete proposed framework
\end{itemize}
Table~\ref{tab:ablation} reveals nuanced performance patterns that illuminate four critical insights regarding component interactions and dataset-specific behaviors:

Table~\ref{tab:ablation} reveals four key insights:
\textbf{Relation-Aware Denoising Effectiveness:} The relation-aware denoising module consistently outperforms alternatives. Replacing it with linear attention yields modest degradation (Recall@20: -0.95\% on Last.FM), while removing explicit denoising causes more significant drops (-2.17\%), confirming its superior noise handling capabilities.
\textbf{Contrastive Learning Impact:} Asymmetric contrastive learning provides consistent improvements over conventional loss, with Beer showing the largest gains (Recall@20: +1.12\%). The effectiveness varies by dataset density, with sparse datasets like Yelp showing minimal sensitivity to contrastive learning variations.
\textbf{Diffusion Augmentation Benefits:} Diffusion-based augmentation primarily enhances ranking quality rather than coverage. It achieves notable NDCG improvements on Last.FM (+1.53\%) with marginal recall gains, suggesting it optimizes embedding discriminability for ranking tasks.
\textbf{Component Complementarity:} Results establish a clear hierarchy: relation-aware denoising dominates recall performance, while diffusion augmentation excels in ranking quality. This demonstrates the complementary nature of components, with denoising enhancing coverage and D-ACL optimizing ranking precision.
\textbf{Dataset-Specific Variance:} On BeerAdvocate, margins among top variants fall within run-to-run variance ($<$0.2\%), while substantial gains on sparse datasets (Last.FM, Yelp) validate the framework's effectiveness in data-scarce scenarios.
\subsection{Model Robustness Test (RQ3)}
In this section, our extensive experimental evaluation demonstrates the efficacy of our proposed RaDAR 
framework. The results indicate that RaDAR exhibits remarkable resilience against data noise and significantly 
outperforms existing methods in handling sparse user-item interaction data. Specifically, our approach maintains 
high performance even in the presence of substantial noise, showcasing its robust nature. 


\begin{figure}[t]
    \centering

    \begin{subfigure}{\linewidth}
        \centering
        \includegraphics[width=0.49\linewidth]{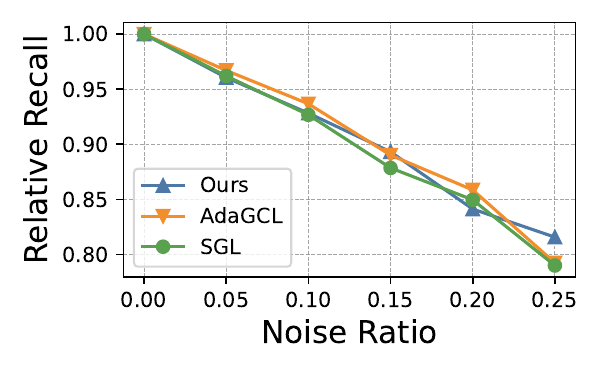}
        \includegraphics[width=0.49\linewidth]{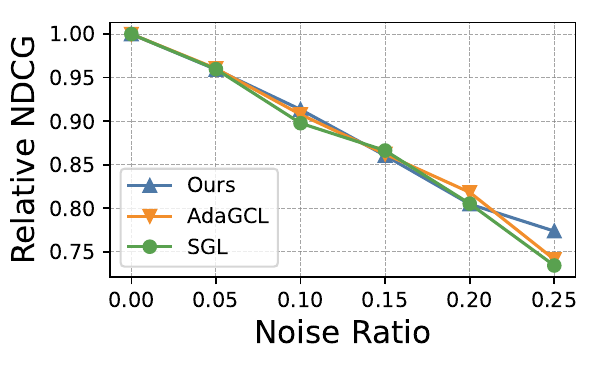}
        \caption{Last.FM data}
    \end{subfigure}

    \vspace{2mm}

    \begin{subfigure}{\linewidth}
        \centering
        \includegraphics[width=0.49\linewidth]{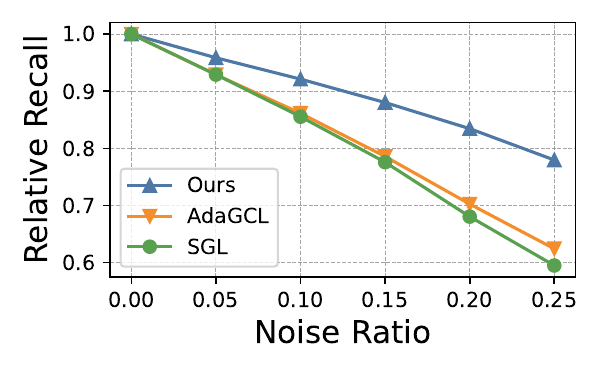}
        \includegraphics[width=0.49\linewidth]{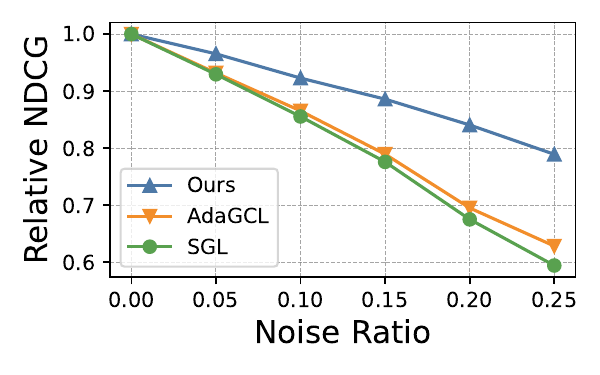}
        \caption{Yelp data}
    \end{subfigure}

    \vspace{2mm}

    \begin{subfigure}{\linewidth}
        \centering
        \includegraphics[width=0.49\linewidth]{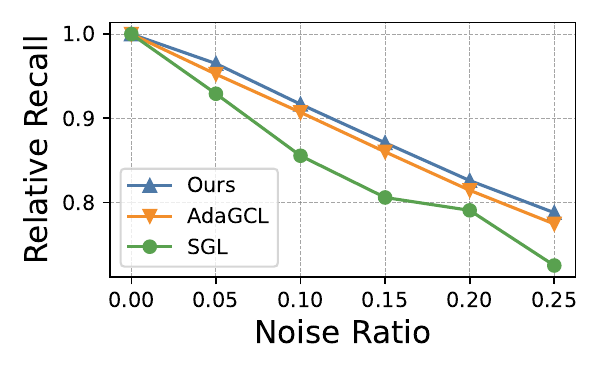}
        \includegraphics[width=0.49\linewidth]{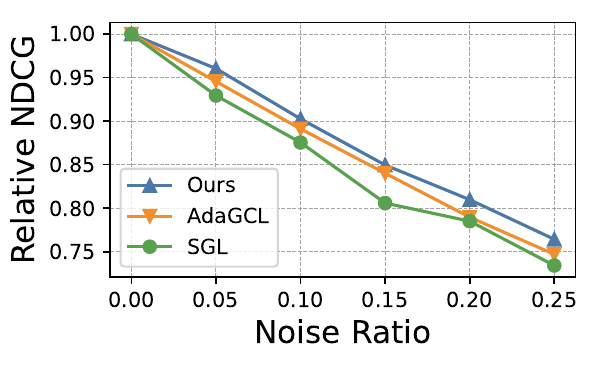}
        \caption{BeerAdvocate data}
    \end{subfigure}

    \caption{Impact of Noise Ratio (5\%--25\%) on Performance Degradation}
    \label{fig:noise_ratio_pic}
\end{figure}

\subsubsection{\textbf{Performance w.r.t. Data Noise Degrees.}}
We systematically evaluate RaDAR's resilience to data corruption through controlled noise injection experiments, where spurious edges replace genuine interactions at incremental ratios (5\%-25\%). A comparative analysis with AdaGCL and SGL across datasets of varying density (Fig.~\ref{fig:noise_ratio_pic}) reveals two key patterns:

On moderate-density datasets (Last.FM: $2.8 \times 10^{-3}$, Beer: $9.5 \times 10^{-3}$), RaDAR demonstrates a modest improvement over AdaGCL on the Beer dataset, while the relative \textit{Recall}/\textit{NDCG} robustness performance among RaDAR, AdaGCL, and GCL shows less significant variation on the Last.FM dataset. This suggests that the benefits of our proposed approach may be less pronounced when data sparsity is moderate, as the existing methods already capture sufficient structural information under these conditions.

In extreme sparsity conditions (Yelp: $1.6 \times 10^{-4}$), RaDAR demonstrates a pronounced advantage with higher relative improvement margins, confirming superior noise resilience in data-scarce scenarios.

Our empirical analysis demonstrates RaDAR's effectiveness in cold-start scenarios through its density-aware denoising framework. The widening performance gap under increasing sparsity highlights the model's ability to extract critical signals from sparse interactions - a pivotal requirement for practical recommendation systems.

\subsubsection{\textbf{Performance w.r.t. Data Sparsity}}
We further examine RaDAR's performance under different levels of user and item sparsity on Yelp. 
From the user perspective (Fig.~\ref{subfig:user_interaction}), RaDAR consistently surpasses AdaGCL across all interaction groups, with the largest improvements observed for cold-start users (0--10 interactions). This confirms RaDAR's robustness in capturing informative user representations even under sparse feedback through adaptive graph augmentation. 
In contrast, the item-side analysis (Fig.~\ref{subfig:item_interaction}) shows that the performance gap widens as item interaction density increases. While user metrics tend to degrade with sparsity (except a mild rebound at 20--25 interactions), item metrics improve steadily with interaction frequency. 
These results demonstrate that RaDAR achieves a balance between user-side generalization under sparsity and item-side learning under dense collaborative signals through its dual adaptive and density-aware mechanisms.

\noindent\textit{Why Recall@20 decreases while NDCG@20 increases under sparsity.}
Interestingly, we observe diverging trends between Recall@20 and NDCG@20 as sparsity increases. Although both metrics are evaluated under the same protocol, Recall@20 tends to drop for highly sparse users, whereas NDCG@20 may increase slightly. This arises mainly from the test set characteristics: (i) many sparse users have only 1--2 held-out positives, where a single miss drastically reduces Recall, but ranking these few positives higher still boosts NDCG; and (ii) on the item side, positives often concentrate on popular items, where improved ranking contributes more to NDCG than to Recall. Hence, under extreme sparsity, the two metrics capture complementary perspectives—Recall reflects coverage, while NDCG emphasizes ranking quality.

\subsection{Hyperparameter Analysis (RQ4)}

We investigate the impact of the adjustable contrastive learning (ACL) ratio $\lambda$, 
which balances Information Bottleneck (IB) losses between the VGAE-base and relation-aware graph denoising view generators. The total IB loss is formulated as $L_{IB} = L_{IB}^G+ \lambda L_{IB}^D$
where $L_{IB}^G$ and $L_{IB}^D$ represent the IB losses from the VGAE-base view generator and the relation-aware graph denoising view generator, and $\lambda > 1$ prioritizes relation-aware structural preservation, while $\lambda < 1$ emphasizes generated graph views.

\begin{figure}[t]
    \centering

    \begin{subfigure}[b]{0.48\linewidth}
        \centering
        \includegraphics[width=\linewidth]{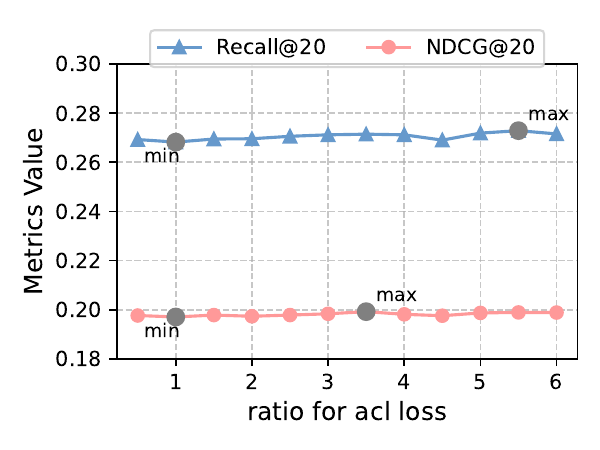}
        \caption{\footnotesize Last.FM}
        \label{subfig:metrics_lastfm}
    \end{subfigure}
    \hfill
    \begin{subfigure}[b]{0.48\linewidth}
        \centering
        \includegraphics[width=\linewidth]{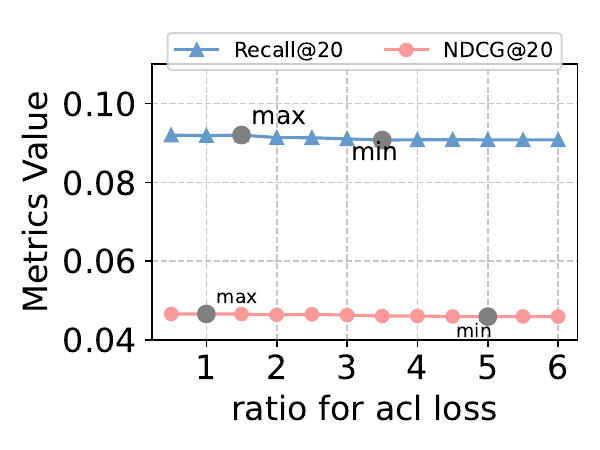}
        \caption{\footnotesize Yelp}
        \label{subfig:metrics_yelp}
    \end{subfigure}

    \caption{Performance variation with ACL ratio $\lambda$. Last.FM peaks Recall@20 at $\lambda=5.5$, NDCG@20 at $\lambda=3.5$. Yelp peaks Recall@20 at $\lambda=1.5$, NDCG@20 at $\lambda=1.0$. Higher $\lambda$ values enhance relation-aware denoising for Last.FM, while Yelp requires balanced contributions due to interaction sparsity.}
    \label{fig:metrics_hyperparameter}
\end{figure}

Fig.~\ref{fig:metrics_hyperparameter} reveals distinct $\lambda$ preferences across datasets. Last.FM achieves optimal performance with 
$\lambda > 1$(Fig.~\ref{subfig:metrics_lastfm}), indicating its structural complexity benefits from enhanced relation-aware denoising. Conversely, Yelp attains peak metrics at lower 
$\lambda$ values (Fig.~\ref{subfig:metrics_yelp}), suggesting its sparse interaction patterns require balanced information preservation from both view generators to prevent overfitting. This empirical evidence confirms RaDAR's adaptability through our symmetric contrastive learning design, showing robust performance across diverse graph recommendation scenarios.

\section{Conclusion}
We present \textbf{RaDAR}, a contrastive recommendation framework that unifies 
(1) generative–denoising dual-view learning, 
(2) asymmetric contrastive objectives, and 
(3) diffusion-based stabilization for noise-robust representation learning. 
RaDAR achieves consistent improvements over state-of-the-art baselines under both binary and weighted-edge regimes, 
demonstrating strong generalization and resilience to sparse or noisy interactions. 
Ablation analyses confirm that diffusion-enhanced denoising and asymmetric contrastive learning jointly contribute to its robustness and stability across datasets.

\section{Limitations.}
RaDAR assumes static graphs without temporal dynamics; extending to sequence-aware diffusion is future work. Training efficiency could be improved via adaptive sampling for large-scale deployment.

\vfill
\bibliographystyle{ACM-Reference-Format}
\balance
\bibliography{ref}

\clearpage
\appendix
\section{Baseline Methods Details}
\label{app:baselines}
\subsubsection{\textbf{Single-behavior Baselines (Binary-edge Setting)}}
We evaluate RaDAR against representative baselines across four research streams covering traditional, neural, graph-based, and contrastive paradigms:
\begin{itemize}[leftmargin=*, itemsep=0.5em]
    \item \textbf{BiasMF}~\cite{koren2009matrix}: A classical matrix factorization model integrating user and item bias terms to enhance personalized preference modeling.
    \item \textbf{NCF}~\cite{he2017neural}: A neural collaborative filtering framework that replaces dot-product interactions with multilayer perceptrons for higher-order user–item relations.
    \item \textbf{AutoR}~\cite{sedhain2015autorec}: An autoencoder-based collaborative filtering method reconstructing user–item interaction matrices for latent representation learning.
    \item \textbf{GCMC}~\cite{berg2017graph}: A graph convolutional autoencoder that models user–item interactions via message passing on bipartite graphs.
    \item \textbf{PinSage}~\cite{ying2018graph}: A graph convolutional framework leveraging random walk sampling and neighborhood aggregation for large-scale recommendation.
    \item \textbf{NGCF}~\cite{wang2019neural}: A neural graph collaborative filtering model capturing high-order connectivity through multi-layer message propagation.
    \item \textbf{LightGCN}~\cite{he2020lightgcn}: A simplified graph convolutional network that removes feature transformations and nonlinearities to enhance efficiency and stability.
    \item \textbf{GCCF}~\cite{chen2020revisiting}, \textbf{STGCN}~\cite{zhang2019star}: Graph-based collaborative filtering variants that refine neighborhood aggregation and address over-smoothing effects.
    \item \textbf{SGL}~\cite{wu2021self}, \textbf{SLRec}~\cite{yao2021self}: Self-supervised graph learning frameworks introducing augmentation and contrastive objectives for robust representation learning.
    \item \textbf{HCCF}~\cite{xia2022hypergraph}, \textbf{SHT}~\cite{xia2022self}: Hypergraph-based self-supervised recommenders capturing both local and global collaborative relations.
    \item \textbf{NCL}~\cite{lin2022improving}: A neighborhood-enriched contrastive learning method that constructs both structural and semantic contrastive pairs.
    \item \textbf{DirectAU}~\cite{wang2022towards}: A representation learning framework optimizing alignment and uniformity on the hypersphere for improved embedding quality.
    \item \textbf{AdaGCL}~\cite{jiang2023adaptive}: An adaptive graph contrastive learning paradigm utilizing trainable view generators for personalized augmentation.
\end{itemize}

\subsubsection{\textbf{Multi-behavior Baselines (Weighted-edge Setting)}}
Following DiffGraph’s weighted-edge protocol~\cite{li2025diffgraph}, RaDAR is further evaluated against baselines specifically designed for multi-behavior or heterogeneous recommendation. To ensure consistency, overlapping models such as BPR~\cite{koren2009matrix}, PinSage~\cite{ying2018graph}, and NGCF~\cite{wang2019neural} are reused but not redundantly described. The remaining representative methods are summarized below:
\begin{itemize}[leftmargin=*, itemsep=0.5em]
    \item \textbf{NMTR}~\cite{gao2019neural}: A neural multi-task model that jointly learns cascading dependencies among multiple user behaviors.
    \item \textbf{MBGCN}~\cite{jin2020multi}: A multi-behavior graph convolutional network disentangling heterogeneous interactions through shared and behavior-specific embeddings.
    \item \textbf{HGT}~\cite{hu2020heterogeneous}: A heterogeneous graph transformer employing type-specific attention and meta-relation projection to model complex interaction semantics.
    \item \textbf{MATN}~\cite{xia2020multiplex}: A memory-augmented transformer network that captures multiplex behavioral relations and long-term dependencies in user behavior.
    \item \textbf{DiffGraph}~\cite{li2025diffgraph}: A diffusion-based heterogeneous graph framework performing noise-aware semantic propagation across relation types for robust link prediction.
    \item \textbf{DDRM}~\cite{wang2023diffusion}: A generative recommendation model built upon denoising diffusion processes, bridging collaborative filtering and generative modeling.
\end{itemize}

\section{Mathematical Details} 
\label{appendix:math_details}
\subsection{Embedding Propagation Details}
\label{appendix:embed_details}
For completeness, we omit duplicated formulas and refer readers to the main text section ``User-item Embedding Propagation'' for the normalized adjacency, layer-wise propagation, final embedding aggregation, and preference scoring formulations.

\subsection{Variational Graph Auto-Encoder Details} 
\label{appendix:vgae_details}
In this section, we provide the detailed mathematical formulations of the VGAE framework used in our view generation approach.
The KL-divergence regularization term for the latent distributions is defined as:
\begin{equation}
\mathcal{L}_{\text{kl}} = -\frac{1}{2} \sum_{d=1}^{D} (1 + 2\log(\mathbf{x}_{\text{std}}) - \mathbf{x}_{\text{mean}}^2 - \mathbf{x}_{\text{std}}^2) 
\label{eq:kl_div}
\end{equation}
For graph structure reconstruction, we employ a discriminative loss $\mathcal{L}_{\text{dis}}$ that evaluates both positive and negative interactions:
\begin{equation}
    \begin{split}
    \mathcal{L}_{\text{pos}} &= \text{BCE}(\sigma(f(\mathbf{x}_{\text{user}}[u] \odot \mathbf{x}_{\text{item}}[i])), \mathbf{1}) \\
    &= -\log(\sigma(f(\mathbf{x}_{\text{user}}[u] \odot \mathbf{x}_{\text{item}}[i]))) \\
    \mathcal{L}_{\text{neg}} &= \text{BCE}(\sigma(f(\mathbf{x}_{\text{user}}[u] \odot \mathbf{x}_{\text{item}}[j])), \mathbf{0}) \\
    &= -\log(1 - \sigma(f(\mathbf{x}_{\text{user}}[u] \odot \mathbf{x}_{\text{item}}[j]))) \\
    \mathcal{L}_{\text{dis}} &= \mathcal{L}_{\text{pos}} + \mathcal{L}_{\text{neg}} 
    \label{eq:vgae_dis}
    \end{split}
\end{equation}
The Bayesian Personalized Ranking (BPR) loss is incorporated to enhance recommendation performance:
\begin{equation}
    \mathcal{L}_{\text{bpr}} = \sum_{(u,i,j)\in O} -\log\sigma(\hat{y}_{ui} - \hat{y}_{uj}), \label{eq:bpr_loss}
 \end{equation}
The total VGAE optimization objective combines these components with weight regularization:
\begin{equation}
    \mathcal{L}_{\text{gen}} = \mathcal{L}_{\text{kl}} + \mathcal{L}_{\text{dis}} + \mathcal{L}_{\text{bpr}}^{\text{gen}} + \lambda_{2}\|\Theta\|_{F}^{2}, \label{eq:gen_loss}
\end{equation}

\subsection{Detailed Diffusion Process Formulation}
\label{app:DetailedDiffusionProcessFormulation}

\subsubsection{Forward Diffusion Process}
\label{sec:ForwardDiffusionProcess}

Our diffusion process begins with the forward phase, where Gaussian noise is progressively added according to:
\begin{equation}
    q(\bm{\chi}_t|\bm{\chi}_{t-1}) = \mathcal{N}(\bm{\chi}_t;\sqrt{1-\beta_t}\bm{\chi}_{t-1}, \beta_t\bm{I})
\end{equation}
with $\beta_t$ controlling the noise scale at step $t$.

The intermediate state $\bm{\chi}_{t}$ can be efficiently computed directly from the initial state $\bm{\chi}_{0}$ through:
\begin{equation}
    \begin{split}
    q(\bm{\chi}_{t}|\bm{\chi}_{0}) = \mathcal{N}(\bm{\chi}_{t};\sqrt{\bar{\alpha}_t}\bm{\chi}_{0}, (1 - \bar{\alpha}_t)\textbf{\emph{I}}), \\
    \bar{\alpha}_t = \prod_{t^{'}=1}^t (1 - \beta_{t^{'}})
    \end{split}
\end{equation}

This allows for the reparameterization:
\begin{equation}
    \bm{\chi}_{t} = \sqrt{\bar{\alpha}_t} \bm{\chi}_{0} + \sqrt{1-\bar{\alpha}_t}\bm{\epsilon}, \bm{\epsilon} \sim \mathcal{N}(0, \textbf{\emph{I}})
\end{equation}

\subsubsection{Linear Noise Scheduler}
\label{sec:LinearNoiseScheduler}

To control the injection of noise in $\bm{\chi}_{1:T}$, we employ a linear noise scheduler that parameterizes $1 - \bar{\alpha}_t$ using three hyperparameters:
\begin{equation}
\begin{split}
    1 - \bar{\alpha}_t &= s \cdot \left[\alpha_{low} + \frac{t-1}{\textit{T}-1}(\alpha_{up} - \alpha_{low})\right], \\
    &t \in \{1,\cdots,\textit{T}\}
\end{split}
\end{equation}

Here, $s \in [0, 1]$ regulates the overall noise scale, while $\alpha_{low} < \alpha_{up} \in (0, 1)$ determines the lower and upper bounds for the injected noise.

\subsubsection{Reverse Denoising Process}
\label{sec:ReverseDenoisingProcess}

The reverse process aims to recover the original representations by progressively denoising $\bm{\chi}_{t}$ to reconstruct $\bm{\chi}_{t-1}$ through a neural network:
\begin{equation}
    p_{\theta}(\bm{\chi}_{t-1}|\bm{\chi}_{t}) = \mathcal{N}(\bm{\chi}_{t-1};\bm{\mu}_\theta(\bm{\chi}_{t},t),\bm{\Sigma}_\theta(\bm{\chi}_{t},t))
\end{equation}
where neural networks parameterized by $\theta$ generate the mean and covariance of the denoising distribution.

\subsection{Asymmetric Contrastive Loss}
\label{sec::acl}
The asymmetric contrastive learning loss function is defined as:
\begin{equation}
\begin{split}
    \mathcal{L}_A &= -\frac{1}{|\mathcal{V}|} \sum_{v \in \mathcal{V}} \frac{1}{|\mathcal{N}(v)|} \sum_{u \in \mathcal{N}(v)} \\
    &\quad \log \frac{\exp(p^\top u / \tau)}{\exp(p^\top u / \tau) + \sum_{v^- \in \mathcal{V}} \exp(v^\top v^- / \tau)},
\end{split}
\label{eq:acl_loss}
\end{equation}
where $\mathcal{N}(v)$ represents the one-hop neighbors of node $v$, and $\tau$ controls the softmax temperature. The predictor output $p = g_{\phi}(v)$ transforms the identity representation into a prediction of its neighborhood context.
\begin{figure}[t]
    \centering

    \begin{subfigure}{\linewidth}
        \centering
        \includegraphics[width=0.49\linewidth]{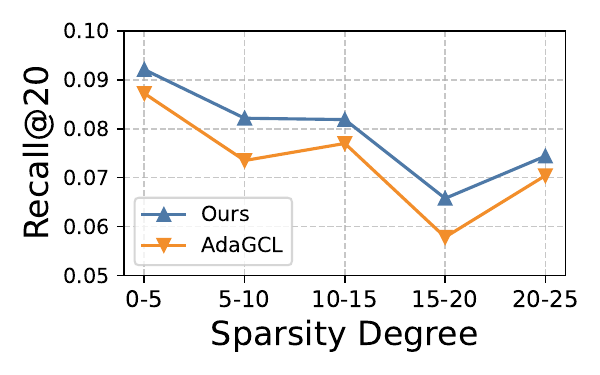}
        \includegraphics[width=0.49\linewidth]{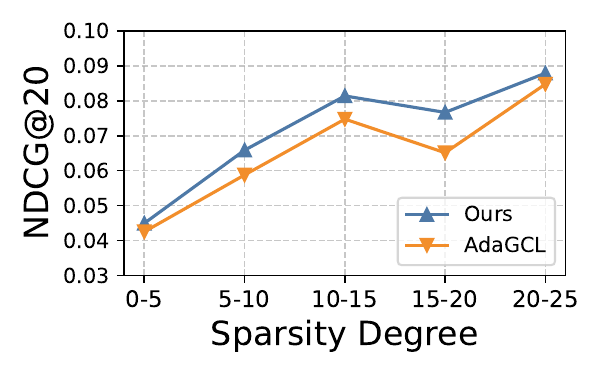}
        \caption{\footnotesize Performance w.r.t. user interaction numbers}
        \label{subfig:user_interaction}
    \end{subfigure}

    \vspace{2mm}

    \begin{subfigure}{\linewidth}
        \centering
        \includegraphics[width=0.49\linewidth]{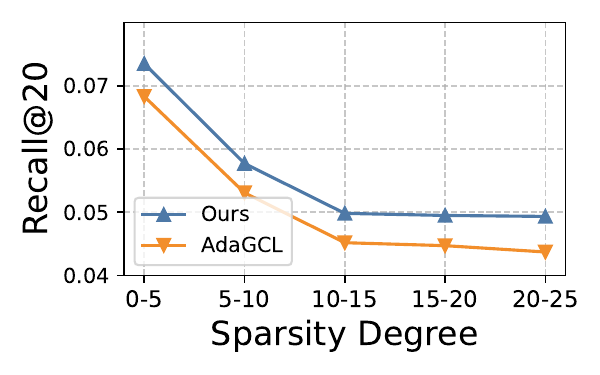}
        \includegraphics[width=0.49\linewidth]{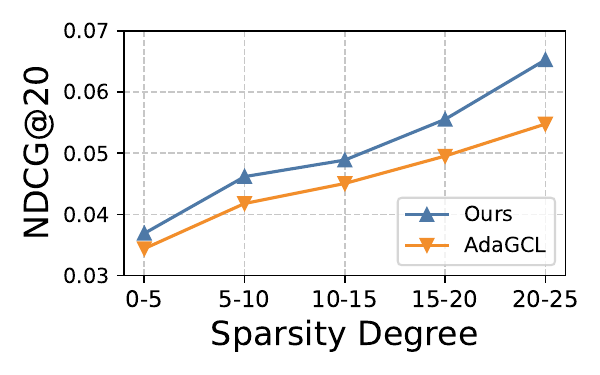}
        \caption{\footnotesize Performance w.r.t. item interaction numbers}
        \label{subfig:item_interaction}
    \end{subfigure}

    \caption{Performance analysis across five user and item interaction sparsity levels on Yelp dataset.}
    \label{fig:interaction_numbers}
\end{figure}
\end{document}